\documentclass[letterpaper, 10 pt, conference]{ieeeconf}  % Comment this line out
\IEEEoverridecommandlockouts                              % This command is only
\usepackage{graphics} % for pdf, bitmapped graphics files
\usepackage{epsfig} % for postscript graphics files
\usepackage{mathptmx} % assumes new font selection scheme installed
\usepackage{times} % assumes new font selection scheme installed
\usepackage{amsmath} % assumes amsmath package installed
\usepackage{amssymb}  % assumes amsmath package installed

\title{\LARGE \bf
Discovering Sequential Patterns in a UK General Practice Database
}

\author{Jenna Reps$^{1}$, Jonathan M. Garibaldi$^{1}$, Uwe Aickelin$^{1}$, Daniele Soria$^{1}$, Jack E. Gibson$^{2}$, Richard B. Hubbard$^{2}$% <-this % stops a space
\thanks{Manuscript received  October 14, 2011.}% <-this % stops a space
\thanks{$^{1}$ School of Computer Science, University of Nottingham, NG8 1BB, UK 
        (phone: +44(0) 11595 14299 ; e-mail:{\tt\small jzr@cs.nott.ac.uk})}
\thanks{$^{2}$ Division of Epidemiology \& Public Health, School of Community Health Sciences, University of Nottingham, NG7 2UH, UK}
}

\begin{document}

\maketitle
\thispagestyle{empty}
\pagestyle{empty}

%%%%%%%%%%%%%%%%%%%%%%%%%%%%%%%%%%%%%%%%%%%%%%%%%%%%%%%%%%%%%%%%%%%%%%%%%%%%%%%%
\begin{abstract}

The wealth of computerised medical information becoming readily available presents the opportunity to examine patterns of illnesses, therapies and responses.  These patterns may be able to predict illnesses that a patient is likely to develop, allowing the implementation of preventative actions.  In this paper sequential rule mining is applied to a General Practice database to find rules involving a patients age, gender and medical history.  By incorporating these rules into current health-care a patient can be highlighted as susceptible to a future illness based on past or current illnesses, gender and year of birth. This knowledge has the ability to greatly improve health-care and reduce health-care costs.

\end{abstract}

%%%%%%%%%%%%%%%%%%%%%%%%%%%%%%%%%%%%%%%%%%%%%%%%%%%%%%%%%%%%%%%%%%%%%%%%%%%%%%%%
\section{INTRODUCTION}

A patient's medical state continuously changes over time, for instance, one day they may be `healthy' and the next they may be `suffering from a cold'. It is common for medical states to develop gradually over time and/or be dependent on previous medical states.  For example, before developing illness A, many patients may previously have illnesses B and C.  If these temporal associations between illnesses can be learned, they can be used to highlight patients that have illnesses B and C as being susceptible to developing illness A.  With this extra knowledge it may be possible to then act to reduce the chance of these susceptible patients developing illness A. Alternatively, if preventing the illness is not possible, susceptible patients may be monitored more frequently to help detect the illness early and improve prognosis.  

Sequential patterning mining algorithms find temporal associations. An example of a sequential pattern rule in the context of retail sales is that 78\% of customers buying an electric toothbrush buy toothbrush replacement heads after three months.  A recent example of mining in a medical context is the application of the sequential pattern mining algorithm FreeSpan on a database known as the RSU Dr. Soetomo medical database to find sequential disease patterns \cite{Yuliana2009}, however, age and gender were not included into the sequential rules and the author only displayed a selection of rules.  Other existing work aiming to detect medical sequential patterns has tended to focus on time series data \cite{Pradhan2009} \cite{Colosimo2002} or specific illnesses, such as investigating patterns that predict the onset of thrombosis \cite{Jensen2001} and identifying traits leading to atherosclerosis in a database of approximately 1400 middle aged men \cite{Klema2008}. There is currently no existing work on detecting sequential patterns of illnesses in The Health Improvement Network (THIN) general practice database (www.thin-uk.com).
%Sequential patterning mining algorithms find temporal associations. An example of a sequential pattern rule in the context of retail sales is that 78\% of customers buying an electric toothbrush buy toothbrush replacement heads after three months.  Existing work aiming to detect medical sequential patterns includes investigating patterns that predict the onset of thrombosis \cite{Jensen2001}, using the sequential pattern mining algorithm freespan on a database known as the RSU Dr. Soetomo medical database to find sequential disease patterns \cite{Yuliana2009} and identifying traits leading to atherosclerosis in a database of approximately 1400 middle aged men \cite{Klema2008}. There is currently no existing work on detecting sequential patterns of illnesses in The Health Improvement Network (THIN) general practice database (www.thin-uk.com).

The aim of this paper is to discover sequential pattern rules occurring within the THIN database. By learning these rules it may be possible to make statements such as `if a patient is born in 1973 and has event A then there is a 70\% chance of them developing event B in the future'. If implemented in health-care monitoring, when event A occurs for any patient born in 1973, then a marker could be added to alert the doctor, potentially allowing preventative actions to be introduced. The advantage of using the THIN database is that the rules learned can be directly incorporated into general practice systems as they contain all the required information, that is, age, gender and medical history.  This may help prevention/early detection of illnesses in many thousands of patients.   
   
The outline of this paper is as follows.  In Section 2 the THIN database and the SPADE algorithm are described. The results of applying SPADE to the THIN database are presented in section 3 followed by a discussion of the importance of the results.  Section 4 presents conclusions and suggests potential future work. 

%For many examples and directions concerning our style, refer to the
%IEEEtran_HOWTO.pdf. The document illustrates many examples of the IEEE style
%and how authors can code their .tex document to comply with this style.
%==================================================================================================================
%   MATERIALS AND METHODS
%==================================================================================================================

\section{MATERIAL AND METHODS} % paste tense as experiements done!
\subsection{THIN Database}
The THIN database contains medical records from participating general practices within the UK.  The data is anonymously extracted directly from the general practice Vision clinical system \cite{inps2011}.  THIN then implements validation steps, these are added as extra fields within the tables.  The database contains patient information including the year of birth, gender, date of registration and family history of each patient registered at the practice since participation. Any information recorded by the doctors when a patient visits (referred to as medical events) is recorded including the date of the visit.  Information regarding any medication prescribed as well as the date of the prescription and the dosage are also included in the database.  In this paper a database containing records from 20 general practices was used.  This subset of the THIN database contained approximately 350 thousand patients, over 25 million prescriptions and over 15 million medical events.

Each medical event is recorded in the database by a reference code known as a Read code. The Read codes used in the THIN database are an independent system designed specifically for primary care but every ICD-9-CM (International Classification of Diseases, Ninth Edition, Clinical Modification) code (or analogues) have a corresponding Read code.

\subsection{Pre-processing Database}
\begin{figure}
\centering
\includegraphics[width=0.5\textwidth]{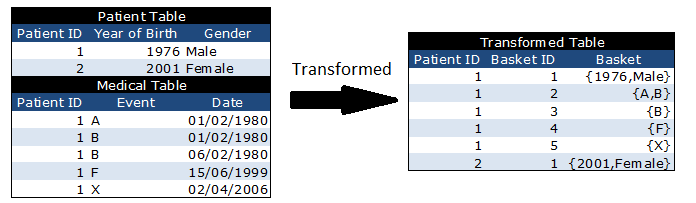}
\centering
\caption{Example of the database transformation.  The left tables are examples of the THIN patient and medical tables for two patients and the right table shows the corresponding transaction database entries for them.}
\label{fig:trans}
\end{figure}

In this paper the THIN database is transformed into a `transaction database', a database commonly used in retail where each entry corresponds to a customer's `basket' (a collection of items purchased during a shopping trip) and each database entry is ordered by the date of the shopping trip.  This transformation is implemented by grouping medical events occurring on the same date for the same patient together into `baskets' and ordering the `baskets' of each patient by the date that the events occurred.  The first `basket' for each patient contains the patient's year of birth and gender.  Medical events with partial or missing transaction dates are ignored from the study as it is not possible to definitively determine their order within the patient's medical sequence. Fig. \ref{fig:trans} shows an example of how the database is transformed for two patients.

%Database transformation example.  The left tables are examples THIN patient and medical tables for patient IDs 1 and 2.  The right table is the transaction database table obtained by transforming the THIN database.
% 
%For each patient their year of birth and gender were entered as the patients first medical transaction.  Medical events for the same patient occuring on the same day were grouped together as a single transaction and the collection of all the patients transactions were ordered by increasing date. An example can be seen in Figure \ref{prepro}.

\subsection{Sequential Pattern Mining}
Sequential pattern mining methods find patterns in ordered sequences. The formal problem is defined as \cite{Agrawal1995}: Given a set of sequential records (sequences) representing a sequential database $D$, a minimum support threshold ($min$\_$sup$) and a set of the unique $items$ $I=\{i_{1}, i_{2},...,i_{k}\}$, find the set of all frequent sequences $S$ in the given sequence database $D$ of items $I$ at the given $min$\_$sup$.

The term $event$ denotes a non-empty unordered collection of items, denoted $(i_{1},i_{2},...,i_{l})$, where each $i_{j}$ is an item.  A $sequence$ is a list of ordered events, denoted $\alpha =  (\alpha_{1} \rightarrow \alpha_{2} \rightarrow ... \rightarrow \alpha_{m}$), where each $\alpha_{j}$ is an event.  The cardinality of a sequence is the number of items it contains and the term $k-sequence$ denotes a sequence of cardinality $k$, $k=\sum_{j}|\alpha_{j}|$.  The sequence $C \rightarrow ABD \rightarrow B$ is an example of a 5-sequence.

A sequence $\alpha = <\alpha_{1} \rightarrow \alpha_{2} \rightarrow... \rightarrow \alpha_{n}>$ is a $sub-sequence$ of another sequence $\beta = <\beta_{1} \rightarrow \beta_{2} \rightarrow... \rightarrow \beta_{m}>$, denoted by $\alpha \sqsubseteq \beta$, if and only if
$\exists i_{1},i_{2},...,i_{n}$, such that $1 \leq i_{1} < i_{2} <...<i_{n} \leq m$ and $\alpha_{1} \subseteq \beta_{i_{1}}, \alpha_{2} \subseteq \beta_{i_{2}},...$ and $\alpha_{n} \subseteq \beta_{i_{n}}$.   

The support of sequence $\alpha$ , denoted sup($\alpha$), is the number of sequences in $D$ that contain $\alpha$, given by 
\begin{equation}
sup(\alpha)= |\{s| s \in D, \alpha \sqsubseteq s\}|
\label{eq:sup}
\end{equation}

The confidence of rule $X \rightarrow Y$ , denoted conf($X \rightarrow Y$ ), is the fraction of sequences containing $X$ that also contain $Y$, as given by
\begin{equation}
conf(X \rightarrow Y)= \frac{sup(X \rightarrow Y)}{sup(X)}
\label{eq:conf}
\end{equation}     

\subsection{SPADE}

Sequential Pattern Discovery using Equivalence Classes (SPADE) \cite{Zaki1998} is a lattice based method and an example of an early pruning algorithm.  These algorithms do not require a support threshold, use position codes and use a vertical projection of the database \cite{Mabroukeh2010}. The SPADE algorithm is suitable for the medical database due to its ability to find sequences with a high confidence even if the items in the sequence are rare.  The reason for this is that SPADE does not require the user to input a minimum value for the support of an item, so rare items (with a low support) can be included into the rules. SPADE was implemented on the pre-processed medical data with a confidence of $0.1$ using the cspade function in the arulesSequences R package \cite{Maindonald2003}, \cite{Buchta2010}.  The confidence value of $0.1$ was chosen for efficiency and to prevent a surplus number of rules being mined.

%==================================================================================================================
%   RESULTS
%==================================================================================================================

\section{RESULTS \& DISCUSSION}

\begin{table}[b]
\centering

\caption{Antecedents of the sequential rules containing Depressive disorder NEC as the consequence and the corresponding confidence values}
\label{tab:dep}

\begin{tabular}{cc}
\hline 
{ Confidence} & { Antecedent} \\ \hline
0.700 & 4x Depressive disorder NEC \\
0.604 & 3x Depressive disorder NEC \\
0.527 & 2x Depressive disorder NEC \\
0.378 & Depressive disorder NEC,Upper respiratory infection \\
0.343& Depressive disorder NEC \\
0.299 & Had a chat to patient,Depressive disorder NEC  \\
0.272 & Depressive disorder NEC,Patient's condition improved \\
0.267 & Depressive disorder NEC,Patient's condition the same \\
0.168 &[D]Insomnia NOS \\
0.138 & Depressed \\
0.106 & Backache  \\
0.105 & 2x Patient's condition improved \\
0.102 & [D]Tiredness\\ \hline 
\end{tabular}
\end{table} 

\begin{table}
\centering
\caption{Antecedents of the sequential rules containing Acute conjunctivitis as the consequence and the corresponding confidence values}
\label{tab:conj}
\begin{tabular}{cc}
\hline 
{ Confidence} & { Antecedent} \\ \hline
0.358& 3x Acute conjunctivitis \\
0.290& 2x Acute conjunctivitis, female \\
0.279& 2x Acute conjunctivitis  \\
0.246& Infect.dis.prevent/control NOS \\
0.227& Acute conjunctivitis,Upper respiratory infection NOS \\
0.224& male,yob:2000 \\
0.223& yob:2003 \\
0.218& Acute conjunctivitis,Otitis media NOS \\
0.215& yob:2001\\
0.203& Acute conjunctivitis,Cough \\
0.193& Acute conjunctivitis,Patient's condition improved\\
0.189& Acute respiratory infection NOS \\
0.185& Acute conjunctivitis,Acute tonsillitis \\
0.182& Acute conjunctivitis,C/O: a rash \\
0.179& Infantile eczema \\
0.178& Croup \\
0.177& Diarrhoea symptom NOS\\
0.175& Non specific viral rash \\
0.173& Sticky eye \\
%0.172& Acute conjunctivitis,Patient's condition the same \\
%0.169& 2x Blood sample $\rightarrow$ Haematol Lab \\
0.167& Molluscum contagiosum \\
%0.167& 2x Urine sample sent to Lab\\
%0.166& Infectious dis:prevent/control,Recurrence of problem \\
%0.166& Stool sample sent to lab. \\
0.165& 3x Acute tonsillitis \\
0.165& Acute non suppurative otitis media \\
0.164& Blood sample (Biochem Lab),Patient's condition same \\
%0.162& Blood sample $\rightarrow$ Haematol Lab,Patient's condition same \\
%\hline
%\end{tabular}
%\caption{Antecedents of the sequential rules containing Acute conjunctivitis as the consequence and the corresponding confidence values}
%\label{tab:conj}
%\end{table} 
%\begin{table}
%\centering
%\begin{tabular}{cc}
%\hline
%{\bf Confidence} & {\bf Antecedent} \\ \hline
%0.158& Allergic conjunctivitis\\
0.158& 2x Acute pharyngitis\\
0.157& Nonsuppurative otitis media + eustachian tube disorders \\
0.155& Refer to Radiology department,female \\
0.154& Suppurative and unspecified otitis media\\
%0.153& Blood sample $\rightarrow$ Biochem Lab, condition improved \\
%0.152& Blood sample $\rightarrow$ Biochem Lab,Refer to Radiology dep \\
0.151& Referral to paediatrician \\
%0.147& Acute tracheitis,female \\
0.145& Blepharitis,female \\
%0.143& O/E - chest examination normal \\
0.140& Normal birth\\
0.138& H/O: viral illness \\
0.137& Enterobiasis - threadworm \\
0.136& Coryza - acute \\
%0.136& Eye symptoms,female \\
%0.133& 2x Atopic dermatitis/eczema\\
%0.131& Child exam.: general/head \\
0.131& Refer to physiotherapist,Patient's condition same \\
0.130& Hordeolum externum ( stye ) \\
0.125& Chickenpox - varicella \\
0.124& Impetigo,female \\
0.123& Verruca plantaris,female \\
0.122& Impetigo \\
0.121& Diarrhoea \& vomiting, symptom \\
0.120& Seborrhoeic dermatitis \\
0.120& Papanicolau smear NEC \\
%0.111& Insect bite NOS\\
0.105& Wheezing \\
0.105& Gastroenteritis \\
%0.104& O/E - dry skin \\
0.102& Adult screening,female\\
\hline 
\end{tabular}
%\vspace{-5mm}
\end{table} 

\begin{table}
\centering
\caption{Antecedents of the sequential rules containing Essential hypertension as the consequence and the corresponding confidence values}
\label{tab:hyper}
\begin{tabular}{cc}
\hline 
{ Confidence} & { Antecedent} \\ \hline
0.344& 2x Essential hypertension,female \\ 
0.333& 2x Essential hypertension\\ 
0.256& High blood pressure \\ 
0.198& O/E - blood pressure reading,female \\ 
0.195& O/E - blood pressure reading \\ 
0.184& Hypertension monitoring,female \\ 
0.177& Hypertension monitoring \\ 
0.174& Essential hypertension,female \\ 
0.166& Essential hypertension \\ 
0.164& Health education offered,Essential hypertension \\ 
0.150& Hypertension monitoring \\ 
0.150& 2x Blood withdrawal\\ 
0.150& 2x Blood sample (Biochem Lab)\\ 
0.132& Chronic kidney disease stage 3 \\ 
0.131& Type 2 diabetes mellitus \\ 
0.127& O/E - blood pressure reading \\ 
0.126& Hypertensive disease,female \\ 
0.122& Hypertensive disease \\ 
0.121& 2x Health education offered \\ 
0.120& ECG \\ 
0.117& yob:1943 \\ 
0.117& Pure hypercholesterolaemia \\ 
0.115& Medication increased\\ 
0.115& Feet examination\\ 
0.114& Blood sample (Biochem Lab),female \\ 
0.111& Gout \\ 
0.110& yob: 1944\\ 
0.109& Hypertension monitoring\\ 
0.108& Blood sample (Lab) NOS\\ 
0.107& Depression screening using questions,female \\ 
0.104& Refer to Radiology department,female \\ 
0.104& Mammography normal \\ 
0.102& Osteoarthritis,female \\ 
0.100& 2x Influenza vaccination\\ 
\hline 
\end{tabular}
\end{table} 

% tells us female/male more likely eg for acute con
% tells us other illnesses seen in the population likely to have consequent event (see below)
% year or birth shows age likely to have illness, conjuct young-most other illness in antecents are 
% young things like chicken pox, threadworm 
% problem due to repeat illness and events entered on same day
% gives info about chance of repeat
% how to tell if illness in antecent is progression or caused 

A total of 97,883 sequential rules were found by SPADE, offering a variety of information. The rules contain information such as on differences between genders, how beneficial health advice is, how age is associated to the illness and other illnesses that may occur while an illness progresses, see Tables \ref{tab:dep} - \ref{tab:hyper}. 

The sequential rules give insight into the number of people who remain ill or relapse.  For example, it was common for the sequential rules to be of the form $A,A \rightarrow A$ or $A,A,A \rightarrow A$, these correspond to patients having the illness again after previously having it two or three times respectively.  The confidence of these sequential rules can be used to estimate the chance of a patient having a repeat illness.  Table \ref{tab:dep} shows 34.3\% of patients suffering from `Depressive disorder NEC' have a repeat but 52.7\% of patients who have `Depressive disorder NEC' twice will have it again, suggesting the probability of them developing `chronic depression' increases each time they have a relapse.  It may be possible to find attributes that increase a patient's chance of relapse by investigating any differences between patients that have repeats and those that do not.  

The difference between the confidences of the sequential rules containing gender information and not containing gender information indicates differences between males and females.  If the sequential rule $A,B,female \rightarrow C$ has a greater confidence than $A,B \rightarrow C$ then this suggests females with a history of events $A$ and $B$ are more likely than males with a history of $A$ and $B$ to develop $C$. For example the rule `Essential hypertension,female'$\rightarrow$`Essential hypertension' has a greater confidence than `Essential hypertension'$\rightarrow$`Essential hypertension', see Table \ref{tab:hyper}, suggesting females are more likely to have a repeat of Essential hypertension than males.  In this way, sequential rules can be used to identify patients that need to be targeted for preventative action.

As the database contains Read codes corresponding to health-care interventions, such as educating and informing patients about an illness, the rules containing these interventions can be analysed to determine if the intervention has been successful. In Table \ref{tab:hyper}, the rule `Health education offered, Essential hypertension' $\rightarrow$  `Essential hypertension' had a lower confidence than `Essential hypertension' $\rightarrow$  `Essential hypertension'.  This suggests a patient may have a decreased chance of repeating the illness if given advice, implying some interventions help improve a patient's medical state.

% talk about age
By adding the patient's year of birth (yob) to the start of each patient sequence it was possible to find age specific illnesses.  The confidence for a rule of the form $yob \rightarrow B$ is the number of patients that were born in $yob$ and had event $B$ by the end of 2010.  This gives some indication of the age that people have event $B$, but with some limitations.  For example, the two rules $1943 \rightarrow$ `Essential hypertension' and $1944 \rightarrow$ `Essential hypertension', Table \ref{tab:hyper}, with respective confidences $0.117$ and $0.110$ show that approximately 11\% of people aged 66/67 in the UK have had `Essential hypertension' at some point. Further, as 1945-2010 was not contained in a sequential rule, a patients chance of having `Essential hypertension' before the age 66 is less than 10\%.  The difference between the confidences for the different $yob$ may help indicate the ages between which there is a rapid increase in patients developing the illness.  For example, patients aged 67 were 6.4\% more likely to have had `Essential hypertension' than patients age 66.  It is strange that $yob=1942$ or lower were not found in any sequential rules, this may be a consequence of the THIN database only having data collected from 2003 onwards. Patients born before 1942 may have progressed from `Essential hypertension'  before the data was recorded.  This also highlights a limitation of the algorithm as it is not able to infer obvious yob rules, such as 'yob less than $1944$'$ \rightarrow$ `Essential hypertension'.   
%only works for old age events as young should find all yobs
      
% talk about other events before progression
For some illnesses that are progressive, sequential pattern mining found events that lead to or cause them. For example, it was found that more than 10\% of patients that have `pure hypercholesterolaemia', `Depression', `Type 2 diabetes mellitus' or `Chronic kidney disease stage 3' develop `Essential hypertension'.  Also, more than 10\% of patients with `Backache' or `Tiredness' were later diagnosed with `depression', see Table \ref{tab:dep}.  Because these rules are common the knowledge is probably well known by doctors. However, the sequential rules provide additional quantitative information. Doctors know that patients with `Type 2 diabetes mellitus' are high risk for developing `Essential hypertension', but the sequential rule confidence gives the actual proportion of patients with `Type 2 diabetes mellitus' that develop `Essential hypertension'.      

% linked events in sequential rules
For some consequences, the antecedents do not lead to or cause the consequence but are linked by the population subgroup with the highest prevalence of the consequent event.  This was observed in Table \ref{tab:conj}, when `acute conjunctivitis' is the consequence.  The $yob$ observed in the antecedents suggests that the young are more susceptible to `acute conjunctivitis'.  Many other antecedents seem to be linked to young age rather than `acute conjunctivitis', such as `Infantile eczema', `Croup', `Normal birth' and `Enterobiasis - threadworm'.  These rules seem most interesting as they are less obvious but may help indicate patients at risk of a future illness by finding illnesses that are common in their population subgroup.     

% talk about un finished sequences, as people are not dead so don't have complete sequence
The limitations with applying sequential pattern mining to all patients in the THIN general practice database is that most patient sequences are not complete and it is difficult to distinguish between real repeat infections or repeat appointments of the same one.  If sequential pattern mining were applied to sequences of patients that have complete records from birth to death, then the set of rules obtained would be complete.  But most patients used in this study are still living and many would only have entries for part of their life recorded.  This may bias the results, as if at age 40 many people develop `event A' and these people are also likely to develop `event B' at age 50 but their subsequence stops at age 42 then, as they are still included for the support and confidence calculations, the support and confidence of the rule `event A' $\rightarrow$ `event B' will be much lower than it should be.  One way of solving this is to only use patients that are registered from birth and also recored as dead, but this may limit numbers.  On the other hand, this may actually help weigh sequential rules as sequential rules that occur over a shorter time interval are less likely to be affected by only having a partial sequence than sequential rules that occur over years, and these shorter interval sequential rules are of greater interest to doctors due to their urgency.          
%==================================================================================================================
%   DISCUSSION
%==================================================================================================================

%\section{Discussion}

%==================================================================================================================
%   CONCLUSIONS
%==================================================================================================================

\section{CONCLUSIONS}
\label{sec:conclusion}
The results obtained in this paper indicate that new information about how medical events, age and gender are related over time can be learned by applying sequential rule mining-algorithms to the THIN longitudinal health-care database by employing the proposed pre-processing methodology.  Interestingly, the key result from this study is that these sequential rules present the possibility of determining the likelihood of re-infections. As the database contains additional information including the demographics of the general practice, family relationships and history, BMI and health adverse activities (eg. smoking), etc, these could  be used to investigate attributes that may increase the chance of a repeat infection. This information offers the potential of preventing re-infection and therefore reducing the cost of health-care in the UK. A comparison between our results and existing results is not possible as existing work either detects patterns in time series data or detects patterns for specific illnesses using different medical attributes than the medical events contained in the THIN database.
%We were unable to compare our results with existing results due to the differences between medical events contained in the databases used in existing work and the THIN database.

Future work needs to address the limitation of discretising the $yob$, so that rules such as `patients born prior to 19xx' $\rightarrow$ `event A' can be mined and to develop a method of identifying if repeat medical entries are repeat entries for the same infection or real reinfections.  It is also of interest to determine the sequential rules obtained when only considering first occurrences of medical events or only using sequences of patients spanning a minimum number of years.      

%%%%%%%%%%%%%%%%%%%%%%%%%%%%%%%%%%%%%%%%%%%%%%%%%%%%%%%%%%%%%%%%%%%%%%%%%%%%%%%%
%\section{ACKNOWLEDGMENTS}

%The authors gratefully acknowledge the contribution of ...

%%%%%%%%%%%%%%%%%%%%%%%%%%%%%%%%%%%%%%%%%%%%%%%%%%%%%%%%%%%%%%%%%%%%%%%%%%%%%%%%

%\bibliographystyle{plain}
%\bibliography{LitRevRef_extras,LitRevRef2}

\begin{thebibliography}{99}

\bibitem{Yuliana2009}
O.Y. Yuliana, S. Rostianingsih and G.S. Budhi, Discovering sequential disease patterns in medical databases using
	FreeSpan mining approach, {\it ICACSIS'09, University of Indonesia, Jakarta, Indonesia}, 2009.	

\bibitem{Pradhan2009}
G.N. Pradhan and B. Prabhakaran, Association rule mining in multiple, multidimensional time series medical data, {\it IEEE international conference on Multimedia and Expo}, 2009, pp. 1716-1719.

\bibitem{Colosimo2002}
A. Colosimo and A. Giuliani and P. Sirabella, Knowledge discovery using medical data mining, {\it Lecture Notes in Computer Science}, 2002, vol. 2526, pp. 1-12.

\bibitem{Jensen2001} 
S. Jensen, Mining medical data for predictive and sequential patterns, {\it PKDD'01, Freiburg, Germany (2001)}, 2001.

\bibitem{Klema2008}
J. Klema, L. Novakova, F. Karel, O. Stepankova and F. Zelezny, Sequential data mining: A comparative case study in development of
	atherosclerosis risk factors, {\it IEEE Transactions on Systems, Man, and Cybernetics: Part C: Applications
	and Reviews}, 2008, vol. 38(1), pp. 3-15.
	
%\bibitem{THIN2010} CSD Medical Research, Cegedim Strategic Data, {\it @online: http://www.epic-uk.org/index.html}, 2011, Visited: 29/07/2011.

\bibitem{inps2011} INPS, A Cegedim Company,Welcome to INPS, {\it @online: http://www.inps4.co.uk}, 2011,  Visited: 30/08/2011.


\bibitem{Agrawal1995}
R. Agrawal and R. Srikant, Mining sequential patterns, {\it Proceedings of the International Conference on Data Engineering}, 1995, pp. 3-14.

\bibitem{Zaki1998}
M.J. Zaki, Efficient enumeration of frequent sequences, {\it Proceedings of the 7th International Conference on Information and
	Knowledge Management.}, 1998, pp. 68-75.

\bibitem{Mabroukeh2010}
N.R. Mabroukeh and C.I. Ezeife, A taxonomy of sequential pattern mining algorithms, {\it ACM Comput. Surv}, 2010,
  vol. 43(1).

\bibitem{Maindonald2003}
J. Maindonald and J. Braun, Data analysis and graphics using R: An example-based approach, {\it Cambridge University Press}, 2003.
  
\bibitem{Buchta2010}
C. Buchta and M. Hahsler, arulesSequences: Mining frequent sequences, {\it @online:http://cran.r-project.org/web/packages/arulesSequences/arulesSequences.pdf}, 2010, Visited: 13/10/2011.


\end{thebibliography}

\end{document}